\documentclass{article}

\usepackage{spconf,amsmath,graphicx}
\usepackage{booktabs}
\usepackage{multirow}
\usepackage{enumitem}
\usepackage{bbding}
\usepackage{amssymb}
\usepackage[cmyk]{xcolor}
\setlist[itemize]{leftmargin=*}
\usepackage{subfigure} 
\usepackage{algorithm,algpseudocode} 

\usepackage{subfig}
\usepackage{subfloat} 
\usepackage{amsthm}
\usepackage{xspace}
\usepackage[normalem]{ulem} 
\useunder{\uline}{\ul}{}

\newcommand{\algoName}{\textsc{UNIDEAL}\xspace}

\newcommand{\w}{\mathbf{w}}

\DeclareMathOperator{\E}{\mathbb{E}}

\newcommand{\norm}[1]{\left\|{#1}\right\|}

\newtheorem{assumption}{Assumption}
\newtheorem{theorem}{Theorem}
\newtheorem{lemma}{Lemma}

\title{\algoName: Curriculum Knowledge Distillation Federated Learning} 
\name{Yuwen Yang$^{\dagger}$, Chang Liu$^{\dagger}$\thanks{$^{\dagger}$Both authors contributed equally to this research.}, Xun Cai, Suizhi Huang, Hongtao Lu, Yue Ding\thanks{$*$Corresponding author: Yue Ding, dingyue@sjtu.edu.cn \\© 2023 IEEE. Personal use of this material is permitted. Permission from IEEE must be obtained for all other uses, in any current or future media, including reprinting/republishing this material for advertising or promotional purposes, creating new collective works, for resale or redistribution to servers or lists, or reuse of any copyrighted component of this work in other works.}$^{*}$}
\address{Department of Computer Science and Engineering, Shanghai Jiao Tong University}

\begin{document}
\ninept
\maketitle
\begin{abstract}
Federated Learning (FL) has emerged as a promising approach to enable collaborative learning among multiple clients while preserving data privacy. However, cross-domain FL tasks, where clients possess data from different domains or distributions, remain a challenging problem due to the inherent heterogeneity. In this paper, we present \algoName, a novel FL algorithm specifically designed to tackle the challenges of cross-domain scenarios and heterogeneous model architectures. The proposed method introduces Adjustable Teacher-Student Mutual Evaluation Curriculum Learning, which significantly enhances the effectiveness of knowledge distillation in FL settings. We conduct extensive experiments on various datasets, comparing \algoName with state-of-the-art baselines. Our results demonstrate that \algoName achieves superior performance in terms of both model accuracy and communication efficiency. Additionally, we provide a convergence analysis of the algorithm, showing a convergence rate of $O(\frac{1}{T})$ under non-convex conditions.
\end{abstract}

\section{Introduction}
\label{sec:intro}
 Federated Learning (FL) is a widely studied distributed machine learning paradigm that enables participants to collaboratively learn a shared model $f$ without collecting data $x$ and output $y$ from local clients~\cite{FedAvg}. However, FL faces the challenge of heterogeneity~\cite{kairouzAdvancesOpenProblems2021a} in real-world scenarios, such as cross-domain caused by cross-region, cross-industry, and other factors (for different clients, data label distribution of $y$ is the same, but the input features of data $x$ are different), and different data collection processes lead to quantity skew. Additionally, varying technical levels or computing capabilities result in the use of different model architectures~\cite{FedMD}. In this work, we focus on addressing the cross-domain problem and further attempt to explore solutions for quantity and model architecture heterogeneity.

Personalized FL (PFL)~\cite{tanPersonalizedFederatedLearning2022} offers multiple paradigms for handling heterogeneity, such as parameter decoupling and knowledge distillation (KD)~\cite{hintonDistillingKnowledgeNeural2015}, which are often employed in situations where model parameters need flexible updates. Some works focus on parameter decoupling by separating model feature extractors and task heads~\cite{FedRep,FeDGen,FedRoD}, channels~\cite{CD2-pFed}, and adapters~\cite{FURL}. For instance, FedRoD~\cite{FedRoD} establishes a dual-head architecture on the client to learn generic representations from different domains and simultaneously improves the global generic and personalized performance of the model by introducing regularization terms. However, many existing works negatively impact the training of cross-domain scenarios, as shown in Table \ref{table:baselines}. We find that sharing task header parameters through PartialAvg yields better performance, as seen in Table \ref{table:ablation}. The KD-based FL method aims to transfer knowledge, such as model parameters~\cite{FedKD}, generative models~\cite{FeDGen}, data embedding~\cite{heGroupKnowledgeTransfer2020,makhijaArchitectureAgnosticFederated2022}, or prototypes~\cite{FedProto,FedPAC,FedPCL}, to guide parameter updating.
FeDGen~\cite{FeDGen} improves the model's generalization across various domains by sharing generators and generating latent space data on each client. However, our experiments reveal that setting inappropriate generators or sharing unsuitable model parameters can adversely affect model performance. FedPAC~\cite{FedPAC} performs refined classifier updates among all clients by sharing global feature representation prototypes and addresses cross-domain issues through feature alignment. However, computing prototypes incurs high computational costs and may expose statistical information of local datasets. KD also typically requires additional datasets~\cite{FedMD}, which complicates dataset preparation and is not conducive to real-world scenarios. In contrast, our approach only requires sharing task header parameters and performing KD on each client's local dataset.

We propose a novel algorithm named \algoName, standing for ``C{\ul u}rriculum K{\ul n}owledge D{\ul i}stillation Fe{\ul de}r{\ul a}ted {\ul L}earning'' by investigating cross-domain scenario.
Building upon the concept of parameter decoupling to address cross-domain inputs and enable further model heterogeneity, our key insight is that such training tasks are particularly challenging at the beginning of the training process. To overcome this difficulty, we employ Curriculum Learning (CL) based KD loss, which encourages clients and the server to find the right training direction through mutual evaluation. By aligning different domains from easy to hard, our approach achieves better convergence and effectively tackles the challenges of cross-domain scenarios.
Results in Sec.\ref{sec:results} show that in the heterogeneous scenario of cross-domain, \algoName achieves the best results in terms of accuracy, communication overhead and running time compared with other SOTA (state-of-the-art) baselines. At the same time, it achieves a convergence rate of $O(\frac{1}{T})$ under non-convex conditions.

\textbf{Contributions.} The contributions of our paper include:
\begin{itemize}[leftmargin=*,noitemsep,topsep=0pt]
    \item  We propose \algoName, a novel FL algorithm designed to address the challenges of cross-domain scenarios and heterogeneous model architectures. It introduces Adjustable Teacher-Student Mutual Evaluation Curriculum Learning, which effectively enhances the knowledge distillation process in FL settings.
    \item We conduct extensive experiments on various datasets, comparing performance of \algoName with state-of-the-art baselines. Results demonstrate that \algoName achieves superior performance in terms of both model accuracy and communication efficiency, highlighting its effectiveness in handling cross-domain FL tasks.
    \item We provide a convergence analysis of \algoName, showing a convergence rate of  $O(\frac{1}{T})$, offering insights into the \algoName's behavior and performance under non-convex conditions.
\end{itemize}

\section{METHODOLOGY} 
To address the challenging cross-domain scenario, we employ a parameter decoupling approach that divides the model parameters into feature extractors and task heads. During the FL process, only the task head parameters are shared. We observe that directly replacing local header parameters with global header parameters is difficult and may lead to performance degradation. Therefore, we propose a curriculum learning-based knowledge distillation method for cross-domain scenarios to facilitate the updating of local header parameters. The details of method are presented in the following sections.

\subsection{Sharing Only Task Head Parameters}
\label{sec:decouple}
Consider a Federated Learning (FL) classification task involving $K$ participating clients. We take a multi-classification task with $C$ classes as an example.  For each client $k\in[K]$, $k$-th client has a model $F(\mathcal{A}_k,\mathbf{w}_k,\cdot)$ with model structure parameter $\mathcal{A}_k$ and model parameter $\mathbf{w}_k = [\mathbf{u}_k,\mathbf{v}_k]$.
Here, $\mathbf{u}_k$ represents the model parameters excluding the header, and $\mathbf{v}_k$ denotes the parameters of the  head.
In client $k$, we maintain two head modules, namely, the global head $\overline{h}$ and local head $h_k$.  At the end of each local training epoch, unlike FedAvg~\cite{FedAvg}, we aggregate only the local head parameters $\mathbf{v}_{k}$ to the server and update the global head's parameter $\overline{\mathbf{v}}$:
\begin{equation}
\small
    \label{eq:aggregation}
    \overline{\mathbf{v}} = \frac{1}{K} \sum_{k=1}^{K} \mathbf{v}_{k}.
\end{equation}

For a sample $\mathbf{x} \in \xi$ in a training batch $\xi$ of size $B$, we can obtain the output batch of global $\overline{\mathbf{y}}$ and local $\mathbf{y}_k$ for client $k$, respectively:
\begin{equation}
\small
    \begin{aligned}
    \overline{\mathbf{y}} = \overline{h} (\overline{\mathbf{v}};f_{\backslash h_k}(\mathcal{A}_k,\mathbf{u}_{k};\mathbf{x})),
    \mathbf{\hat{y}}_k = h_k (\mathbf{v}_{k};f_{\backslash h_k}(\mathcal{A}_k,\mathbf{u}_{k};\mathbf{x})),
    \end{aligned}
\end{equation}
where $h_k$ and $f_{\backslash h_k}$ denote the head and non-head modules, respectively.
This parameter decoupling enables FL to disregard the content before the head module $h$, thereby accommodating the heterogeneity of input data and the majority of models.

There are two significant advantages to using parameter decoupling and sharing only the task head parameters. First, it requires the transfer of only a small number of model parameters. For example, when using ResNet-18~\cite{resnet} for the image dataset in our experiments, only approximately 2.17\% (0.11MB of 5.08MB) of the parameters' size needs to be transferred. This greatly reduces communication overheads and privacy leak risks. Second, Table \ref{table:baselines} and \ref{table:ablation} demonstrate that, in heterogeneous cross-domain scenarios, transmitting all model parameters or feature extractor parameters can easily result in performance degradation. In contrast, sharing only the task head parameters can help improve the model's performance.

\subsection{Adjustable Teacher-Student Mutual Evaluation Curriculum Learning}
\label{sec:CLKD}
Parameter decoupling enables FL among clients with different features. However, the cross-domain features hinder the model's convergence. To address this issue, we leverage the idea of curriculum learning \cite{CLSurvey} to gradually align domains ``from easy to hard.''

Specifically, we employ knowledge distillation as a constraint to align different domains. Following the parameter decoupling in Sec.\ref{sec:decouple}, we consider the global readout module $\overline{h}$ as a teacher capable of handling cross-domain data. However, the teacher $\overline{h}$'s  exhibits poor performance at the beginning of training. As a result, it is impractical to directly perform knowledge distillation between teacher $\overline{h}$ and student $h_k$ on all training samples for client $k$. 

We propose an ``Adjustable Mutual Evaluation Teacher-Student \textbf{C}urriculum \textbf{L}earning for \textbf{K}nowledge \textbf{D}istillation'' method, dubbed as \textbf{CLKD}, to enhance training. Intuitively, the more similar the teacher and student outputs are, the easier the samples for knowledge distillation. We compute the similarity between the output of teacher $\overline{\mathbf{y}}$ and student $\mathbf{y_k}$ using a specific metric as the teacher-student mutual evaluation score $\mathbf{s} \in R^{ B}$ for the training batch samples.
\begin{equation}
\small
    \label{eq:score}
    \mathbf{s} = \text{Metric}(\overline{\mathbf{y}}, \mathbf{\hat{y}}_k),\text{Metric}_\text{cos}(\overline{\mathbf{y}}_i, \mathbf{\hat{y}}_{k,i})=\frac{\overline{\mathbf{y}}_i\cdot \mathbf{\hat{y}}_{k,i}}{\|\overline{\mathbf{y}}_i\|\times \|\mathbf{\hat{y}}_{k,i}\|},
\end{equation}
where Metric($\cdot,\cdot$) can be any measure that quantifies the similarity between two samples. Here, $i \in [B]$ denotes the sample index of the training batch, and $\|\cdot\|$ represents the $L_2$ norm of the input vector. We also compare the reciprocal of $L_1$ and $L_2$ norm in our experiments and find that cosine similarity is more suitable for \algoName. Please refer to Sec. \ref{table:ablation} for details.

Instead of manually setting the curriculum learning threshold, we implement an adjustable threshold $s_T$ for each training batch. Given the relative proportion $p$ used for training, we use the rounded $p\times B$-th descending sorted teacher-student mutual evaluation score as the threshold $s_T$:
\begin{equation}
\small
\label{eq:threshold}
    s_T = \text{Sort}(\mathbf{s})[{\text{Round}(p\times B)}],
\end{equation}
where $\text{Sort}(\cdot)$ is a descending sorting function and $\text{Round}(\cdot)$ is a rounding function. 
For a given proportion $p$, we train models with the $(1-p)\times B$ easiest samples. In other words, we only compute the knowledge distillation loss for samples ``easier than the threshold $s_T$'', as follows:
\begin{equation}
\small
\label{eq:CLloss}
    \mathcal{L}_{\text{CL},k}(\overline{\mathbf{y}}, \mathbf{\hat{y}}_k, s_T) = \sum_{e=0}^{E-1} \sum_{i=0}^{B-1} \text{KL}(\sigma(\overline{\mathbf{y}}^{n}_i), \sigma(\mathbf{\hat{y}}^{n}_{k,i})) \mathbb{I}(\mathbf{s}^{n}_i \geq s^{n}_T),
\end{equation}
where the $\mathcal{L}_{\text{CL},k}$ is the CLKD loss and $E$ is the number of batch for $k$-th client, the adjustable threshold $s_T$ is computed by Eq.(\ref{eq:threshold}), $\text{KL}(\cdot)$ is the Kullback-Leibler Divergence, $\sigma(\cdot)$ is the softmax function, $\mathbb{I}(\cdot)$ is the indicator function, and the teacher-student mutual-evaluation score $\mathbf{s}$ is computed by Eq.(\ref{eq:score}).
Combining all the above formulas, our personalized objective function takes the following form:
\begin{equation}
\label{eq:final_opt}
\begin{aligned}
\min _{\{\mathbf{w}_k\}} \frac{1}{K} \sum_{k=1}^K \mathcal{L}_k(\mathbf{w}_k) &= \frac{1}{K} \sum_{k=1}^K [ \mathcal{L}_{\text{CE},k}(\mathbf{w}_k) + \frac{\alpha}{2} \mathcal{L}_{\text{CL},k}(\mathbf{w}_k) ],\\
\end{aligned}
\end{equation}
where $\mathcal{L}_{\text{CE}}$ is the empirical loss and hyperparameter $\alpha$ represents the absolute strength of CLKD Loss. To enable curriculum learning to train model from the easy to hard samples, we set the proportion $p$ to decrease from $\frac{1}{B}$ to zero linearly. In other words, it computes the $\mathcal{L}_{\text{CL}}$ with only one easiest sample of each training batch for the first epoch, gradually adding harder samples until all samples can be trained for the last epoch. The above algorithm is designed for single-round communication. In each round, clients use gradient descent to update the model parameters according to formula \eqref{eq:final_opt}.

\subsection{Extension for Heterogeneous Architecture Models}
In addition to addressing data heterogeneity, \algoName can also handle FL challenges with heterogeneous model architectures $\mathcal{A}_k$ , as it only requires transmission of header parameters during communication process. On one hand, this enables each participant to fully develop feature extractors for their own datasets, thereby improving the effectiveness of local models. On the other hand, the introduction of heterogeneous models further reduces the possibility of data being reverse-engineered, a common issue in scenarios with fully shared model parameters~\cite{wang2020attack}, and better preserves user information privacy and security. Results in Tables \ref{table:baselines} and \ref{table:ablation} also demonstrate that \algoName-HETE benefits from the heterogeneous model architecture and achieves improved model performance.

\begin{table*}[htbp]
\centering
\resizebox{\linewidth}{!}{
\begin{tabular}{@{}lcccccccc|ccccc@{}}
\toprule
\multicolumn{1}{c}{\multirow{2}{*}{Method}} & \multicolumn{2}{c}{DIGIT} & \multicolumn{2}{c}{ADULT} & \multicolumn{2}{c}{HCC} & \multicolumn{2}{c}{ILPD}
&\multicolumn{1}{c}{\multirow{2}{*}{$C_r$}} &\multicolumn{1}{c}{\multirow{2}{*}{$T_r$}}&\multicolumn{1}{c}{\multirow{2}{*}{$N_c$}}&\multicolumn{1}{c}{\multirow{2}{*}{$C_t$}}&\multicolumn{1}{c}{\multirow{2}{*}{$T_t$}}\\
\multicolumn{1}{c}{} & NIID-1     & NIID-2     & NIID-1     & NIID-2     & NIID-1     & NIID-2     & NIID-1     & NIID-2     \\ \midrule
Local                & 94.36 & 94.44 & 80.32 & 75.80 & 85.13 & 78.31 & 76.39 & 74.62 & 0.00  & 7.34  & 4  & 0.00  & 29.36\\ \midrule
FedAvg~\cite{FedAvg} & 93.25 & 91.12 & 80.15 & 74.74 & 81.86 & 71.14 & 78.40 & 75.82 & 5.08  & 7.52  & 5  & 25.38 & 37.60\\
FedRep~\cite{FedRep} & 92.73 & 92.11 & 80.22 & 75.88 & 84.04 & 77.56 & 77.36 & 76.27 & 5.06  & 8.16  & 10 & 50.61 & 81.60\\
FedBABU~\cite{FedBABU} & 91.92 & 90.96 & 79.90 & 74.93 & 81.54 & 76.53 & 76.53 & 74.93 & 5.06  & 9.38  & 8  & 40.49 & 75.04\\
FedRoD~\cite{FedRoD}   & 93.20 & 92.99 & 79.87 & 75.43 & {\ul 85.13} & {\ul 79.41} & 75.76 & 74.32 & 5.08  & 7.71  & 6  & 30.45 & 46.26\\
FedProto~\cite{FedProto} & 90.46 & 91.09 & 79.68 & 76.78 & 82.37 & 69.74 & 48.47 & 52.08 & 0.01  & 20.65 & 6  & 0.09  & 123.90\\
FedKD~\cite{FedKD}     & {\ul 94.57} & {\ul 94.84} & 79.82 & 75.61 & 82.44 & 75.35 & 78.82 & 76.85  & 4.58  & 19.62 & 6  & 27.48 & 117.72\\
FedPCL~\cite{FedPCL}   & 24.36 & 24.22 & 78.48 & 76.91 & 82.63 & 71.35 & 46.47 & 57.29 & 0.01  & 15.90 & -     & -     & -\\
FedPAC\cite{FedPAC}    & 93.77 & 93.72 & {\ul 80.23} & {\ul 79.46} & 83.21 & 74.23 & {\ul 79.17} & {\ul 79.38} & 5.09  & 9.71  & 8  & 40.72 & 77.68\\
FeDGen~\cite{FeDGen}   & 29.72 & 30.34 & 76.98 & 71.47 & 62.56 & 63.03 & 73.33 & 76.12 & 10.34 & 13.40 & -     & -     & -\\
FeDGen-P~\cite{FeDGen} & 35.85 & 46.48 & 77.52 & 66.07 & 56.79 & 61.42 & 76.26 & 70.85 & 5.37  & 13.01 & -     & -     & - \\ \midrule
\algoName              & 94.88 & 94.69 & 80.39 & 74.94 & \textbf{85.96} & 87.96 & 79.50 & \textbf{83.76} & 0.11  & 7.48  & 2  & 0.22  & 14.96\\
\algoName-HETE        & \textbf{95.38}  &\textbf{ 95.87}  & \textbf{80.41}  & \textbf{79.88}  & 85.38 & \textbf{88.12} & \textbf{79.55}  & 80.90 & 0.11  & 7.42  & 2  & 0.22  & 14.84\\ \midrule
$p$-value              & 2.00E-03   & 4.30E-06   & 3.93E-03   & 7.39E-03   & 1.35E-01   & 4.24E-09   & 4.98E-01   & 1.29E-07   \\ \bottomrule
\end{tabular}
}
\caption{Experiments show \algoName outperforms these state-of-the-art methods. The left table: \textbf{Top-1 test accuracy comparisons} (mean on 10 trials, \%) on different heterogeneous settings and datasets.  
-HETE means using heterogeneous model architecture. \textbf{Bold} denotes the highest accuracy. {\ul Underline} denotes the highest accuracy of baselines. $p$-value reports the Student’s T-Test between the aforementioned two results. The right table: \textbf{Communication overhead and running time comparison in DIGIT-NIID-1 setting.} $C_r$ and $T_r$ are the communication overhead (MB) and running time (s) for each round respectively, $N_c$ is number of rounds required to achieve 50\% accuracy, $C_t$ and $T_t$ are the corresponding total communication overhead and total time respectively.
} 
\label{table:baselines}
\end{table*}

\begin{table*}[htbp]
\vspace{-0.2cm}
\centering
\setlength{\abovecaptionskip}{0pt}
\resizebox{\linewidth}{!}{
\begin{tabular}{@{}lccccccccc@{}}
\toprule
\multirow{2}{*}{Method} &  \multirow{2}{*}{Local} &  \multirow{2}{*}{FedAvg} &  \multirow{2}{*}{FedRep} &  \multirow{2}{*}{PartialAvg} &  PartialAvg &
\multirow{2}{*}{PartialKD} &  PartialKD &  PartialKD &  PartialKD \\
     &            &            &            &            & w/PartialKD &            & w/$L_1$      & w/$L_2$      & w/$cos$  \\ \midrule
HOMO & 94.36±0.22 & 93.25±0.51 & 92.73±0.51 & 94.36±0.17 & 94.39±0.17   & 94.79±0.25 & 94.85±0.19 & 23.79±5.83 & 94.88±0.17 \\
HETE & 94.92±0.53 & -          & -          & 94.94±0.35 & 94.63±0.71   & 94.99±0.46 & 95.22±0.57 & 95.17±0.36 & 95.38±0.56 \\ \bottomrule
\end{tabular}
}
\caption{Ablation experiments in the DIGIT-NIID-1 setting. HOMO stands for homogeneous model architecture settings and HETE heterogeneous model architecture settings.  w/ stands for ``with''. $L_1$, $L_2$ and $\text{cos}$ respectively represent the use of $L_1$, $L_2$ norm and cosine similarity to calculate metric $s$ in equation \eqref{eq:score} for CLKD loss. Note that the PartialKD with $\text{Metric}_\text{cos}$ in the last column is our \algoName algorithm.}
\label{table:ablation}
\end{table*}

\begin{figure}[htbp]
\setlength{\abovecaptionskip}{0pt}
\setlength{\abovedisplayskip}{0pt}
\centering
\includegraphics[width=0.9\linewidth]{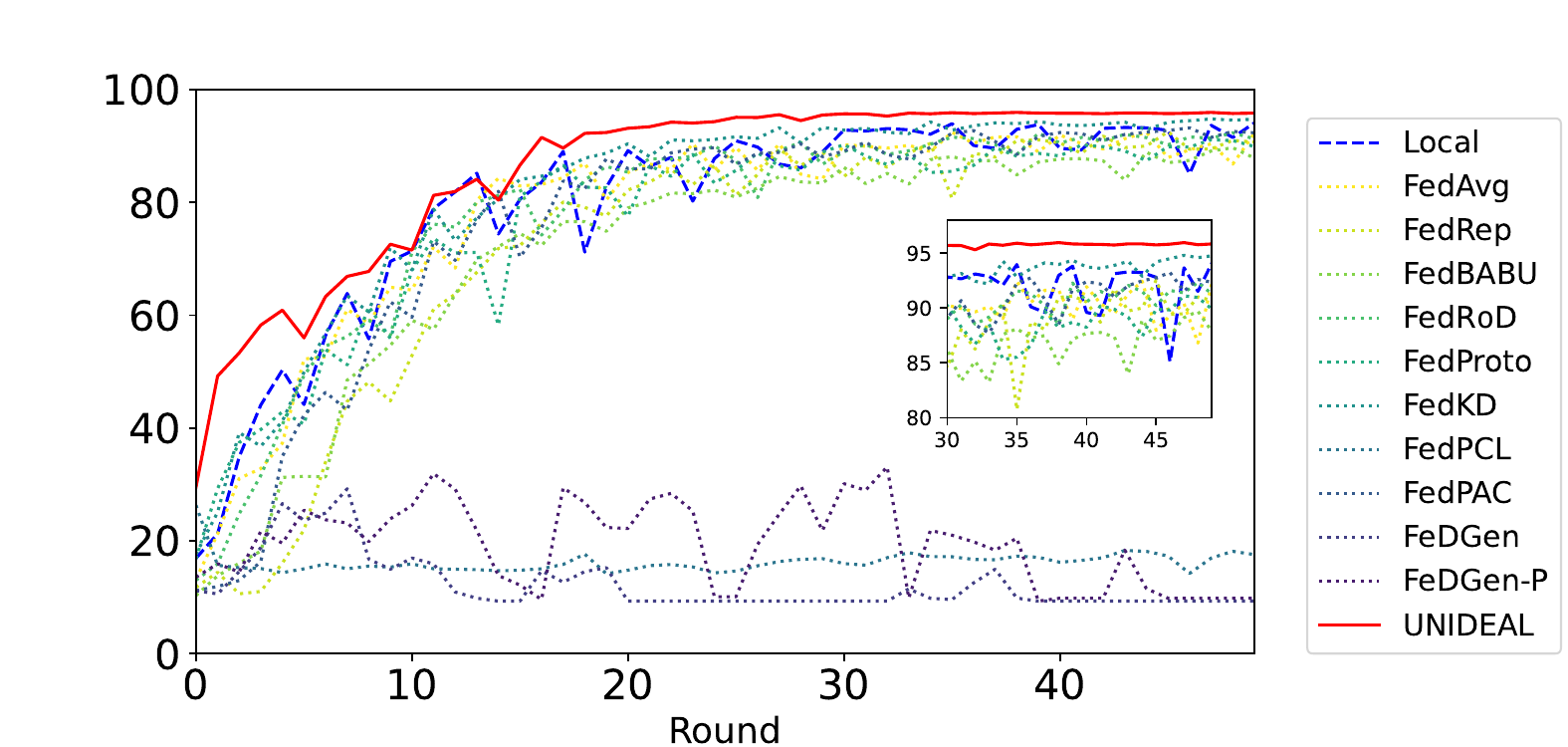}
\caption{Test accuracy varies with communication rounds in the DIGIT-NIID-1 setting. \algoName achieves better accuracy improvement with fewer rounds, and reaches higher accuracy than other baselines in the later stage while maintaining stable accuracy.}
\label{fig:acc}
\end{figure}

\section{Experiments}
\subsection{Experimental Setup}
\textbf{Image datasets and models.} To simulate cross-domain scenarios, we introduce MNIST \cite{MNIST}, MNIST-M ($1\times28\times28$ gray-scale images) \cite{MNIST-M} ($3\times32\times32$ RGB images), and Synthetic Digits ($3\times32\times32$ RGB images) \cite{MNIST-M} as our test datasets. We choose ResNet-18 \cite{resnet} as the backbone in homogeneous settings and add MobileNet-V3 \cite{mobilenet} and VGG \cite{vgg} for the second and third client respectively in heterogeneous setting. 

\textbf{Tabular datasets and models.} We select the ADULT \cite{UCI}, HCC \cite{HCC}, and ILPD \cite{UCI} datasets to evaluate the method's performance in tabular scenarios. First, we slice each dataset into sub-datasets with different features and sample IDs and distribute them to each client. Then, we divide the training and testing sets within each client. In homogeneous settings, we use the same MLP model, while in heterogeneous model architecture settings, we generate MLP models randomly with different layers and hidden unit numbers for each client.

\textbf{Baselines.} In addition to only local training (Local), we choose ten state-of-the-art (SOTA) baselines for comparative experiments. (a) FedAvg~\cite{FedAvg} is a pioneering FL method; (b) FedRep~\cite{FedRep}, FedBABU~\cite{FedBABU}, and FedRoD~\cite{FedRoD} focus on parameter decoupling; (c) FedProto~\cite{FedProto}, FedKD~\cite{FedKD}, FedPCL~\cite{FedPCL}, FedPAC~\cite{FedPAC}, and FeDGen~\cite{FeDGen} employ knowledge distillation with model parameters or prototypes. FeDGen-P is an expanded version of FeDGen that shares model header parameters. In ablation experiments, PartialAvg performs FedAvg~\cite{FedAvg} operations only on the model head parameters with (\ref{eq:aggregation}) during training. PartialKD uses global model head parameters to guide local head parameter updates through knowledge distillation. Adding CLKD loss \eqref{eq:CLloss} based on cosine similarity to PartialKD results in \algoName.

\textbf{Implementation Details.} We develop the algorithm using PyTorch and implement comparative experiments on top of PFL Platform~\cite{FedALA}. All methods are trained on a single RTX 2080Ti GPU. 

\textbf{FL setting.} In NIID-1, we use different datasets on all clients for image datasets and randomly sample half of the features of each dataset on all clients for tabular datasets to achieve cross-domain settings. We also use a Dirichlet distribution (alpha is 0.5) to redistribute sample labels in the NIID-1 setting to create a more heterogeneous setting, NIID-2.

\textbf{Parameter Setting.} We record the best average test accuracy of all clients each time. We repeat each experiment with ten continuous random seeds and compute the mean and standard deviation of accuracy without cherry-picking. The maximum number of communication rounds, $T_{max}$, is 50. To further ensure a fair comparison, we use the same number of training rounds and adjust the hyperparameters (such as learning rate and batch size) to ensure fully converges. The strength $\alpha$ of CLKD in \eqref{eq:final_opt} is 1. Other unspecified parameters remain appropriate and consistent across each algorithm.

\subsection{Main Results}
\label{sec:results}
\textbf{Overall comparison.} From the comparative experimental results in Table \ref{table:baselines}, \algoName has achieved significantly better performance than other baselines (at the significance level $p$-value = 0.01; in eight datasets compared with ten baselines, it achieves significant advantages in 78 groups of situations and achieves consistent optimal performance with the best-performing baseline algorithm in two situations). As can be seen from Table \ref{table:baselines} and Fig. \ref{fig:acc}, when reaching the same accuracy of 50\%, the communication overhead and running time of our algorithm are the least, and the communication overhead and running time per round are also in the first echelon.

\textbf{Cross-domain remains a challenging FL scenario.} Most of the algorithms in Table \ref{table:baselines} perform worse than the Local training. This demonstrates that in heterogeneous scenarios, inappropriate FL methods can hinder the model training process.

\textbf{Parameter decoupling at the right location and a softer parameter update approach are better suited to cope with cross-domain FL scenarios.} As observed from Table \ref{table:ablation}: (1) compared with Local training, FedRep, which only transmits the model feature extractor parameters, results in a greater performance drop than FedAvg, which transmits all model parameters. However, PartialAvg, which only transmits model header parameters, achieves consistent performance with Local. (2) PartialKD, which uses global header parameters to guide model parameter updates only through knowledge distillation, performs better than the other two PartialAvg algorithms that directly replace local header parameters at each round.

\textbf{Adjustable Teacher-Student Mutual Evaluation Curriculum Learning can further improve the effectiveness of knowledge distillation.} Table \ref{table:ablation} shows that CLKD based on cosine similarity provides more significant performance improvements than CLKD based on the $L_1$ norm. Considering the results from all homogeneous and heterogeneous model architecture scenarios, we choose cosine similarity as the metric for CLKD.
CLKD not only achieves the best performance among all methods considered but also maintains low running time, outperforming all competitors. This demonstrates the effectiveness of the Adjustable Teacher-Student Mutual Evaluation Curriculum Learning approach in balancing both model performance and communication efficiency, making it a superior solution for cross-domain scenarios.

\textbf{UNIDEAL can gain training benefits from carefully designed models by participants.} In experiments on the DIGIT dataset of Tables \ref{table:baselines} and \ref{table:ablation}, we selected the best locally performing model on the three clients for the UNIDEAL-HETE algorithm. The final model results show that in heterogeneous model architecture FL scenarios, UNIDEAL-HETE can indeed achieve performance improvements consistent with Local(HETE), demonstrating the effectiveness of carefully designed models in enhancing FL outcomes.

\section{Convergence Analysis}

Here, we provide a brief version of the algorithm convergence analysis. The model architecture parameters $\mathcal{A}_k$, which are used to determine the heterogeneous network of each client, are not updated according to the gradient, so they can be ignored in the convergence analysis. We treat the $k$-th model's parameters $\mathbf{w}_k=[\mathbf{v}_{k};\mathbf{u}_{k}]$ as a whole. We make the same assumptions about the model parameters as \textbf{Assumption 1 in FedSSD}~\cite{FedSSD} (Bounded dissimilarity, L-Lipschitz smooth, and L-Lipschitz continuity) and \textbf{Assumption 1.3 in FedGKD}~\cite{FedGKD}.
Similar to FedGKD, we also construct $\gamma$-inexact solutions using \textbf{Definition 2 in FedProx}~\cite{FedProx}.

\begin{lemma}\label{lemma:wellposed}
Define $\mathcal{\tilde{L}}$ as follows:
\begin{equation}
\small
\label{eq:lemma1}
\begin{aligned}
 \mathcal{L}(\w;\w^{t}) 
 &= \mathcal{L}_{\text{CE}}(\w) + \frac{\alpha}{2} \mathcal{L}_{\text{CL}}(\mathbf{w}), \\ 
  & \leq \mathcal{L}_{\text{CE}}(\w)+  \frac{\alpha L_h }{2\delta}\norm{\w^t - \w }^2\overset{\Delta}{=} \mathcal{\tilde{L}}(\w;\w^{t}),
\end{aligned}
\end{equation}
where the equation is from the definitions of loss function \eqref{eq:final_opt} and inequality is because the indicator function in \eqref{eq:CLloss} is less than or equal to 1 and Assumption 1.3 in FedGKD~\cite{FedGKD}. 
Notice that for any approximate solution $\w_k^{t+1}$ satisfies $\tilde{\mathcal{L}}(\w_k^{t+1};\w^{t}) \leq \tilde{\mathcal{L}}(\w^t;\w^{t})$, then $\mathcal{L}(\w_k^{t+1};\w^t) \leq \tilde{\mathcal{L}}(\w_k^{t+1};\w^{t}) \leq \tilde{\mathcal{L}}(\w^t;\w^{t}) = \mathcal{L}(\w^t;\w^t)$
which implies that a solution optimizing $\tilde{\mathcal{L}}$ also satisfied $\mathcal{L}$.
\end{lemma}

\begin{assumption}\label{ass.algo}
  At the $t$-th round, the $k$-th client solves the optimization problem  $\w_{k}^{t+1} \approx\arg\min_{\w}\;L_k(\w;\w^{t})$ approximately, satisfies: $
    \norm{\nabla \mathcal{L}_{\text{CE},k}(\w_{k}^{t+1}) + \frac{\alpha L_h }{\delta}(\w_k^{t+1}-\w^t)}\leq \gamma \norm{\nabla \mathcal{L}_{\text{CE,k}}(\w^{t})}\,,$
where $\gamma\in [0,1)$.
\end{assumption}
Lemma~\ref{lemma:wellposed} shows the well-posedness of the Assumption~\ref{ass.algo}. 

\begin{theorem}[Convergence]\label{thm:convergence}
Let Assumptions above hold. Assume for each round, a subset of $S_t$ clients are selected, with $|S_t|=K$ and the $k$th client is selected with the probability $p_k$. If $\lambda_{\min}, L$ and $B$ satisfy Assumption 1 in FedSSD~\cite{FedSSD}. And if $\gamma$ and $K$ are chosen to satisfy:
\begin{enumerate}
\small
    \item $\overline{\mu} := \frac{\alpha L_h}{\delta} + \lambda_{\min}>0$;
    \item $\rho>0$, where $\rho = \frac{\delta}{L_h}\left(1-\gamma B-\frac{LB(1+\gamma)}{\overline{\mu}}\right) \\
          - \frac{1}{\overline{\mu}}\left(\frac{\sqrt{2}B(1+\gamma)}{\sqrt{K}} + \frac{L(1+\gamma)^2B^2}{2\overline{\mu}} \right.
         + \left. \frac{(2\sqrt{2K}+2)LB^2(1+\gamma)^2}{\overline{\mu} K}\right)$
\end{enumerate}
then after $t$ rounds, $\E_{S_t}[\mathcal{L}(\w^{t+1})] \leq \mathcal{L}(\w^t) - \rho \norm{\nabla \mathcal{L}(\w^t)}^2$.
\end{theorem}
For the proof of Theorem \ref{thm:convergence}, please refer to Theorem 4 in FedProx\cite{FedProx}. Next, we take the total expectation of the inequality above with respect to all randomness (including local stochastic gradient descent and client sampling) on both sides. Finally, we divide both sides by $T$ to obtain the desired result:
\begin{equation}
\small
\min_{t\in[T]} \E[\norm{\nabla \mathcal{\tilde{L}}(\w^t)}]\leq \frac{1}{T}\sum_{t=0}^{T-1}\E[\norm{\nabla \mathcal{\tilde{L}}(\w^t)}]^2  \leq \frac{\mathcal{\tilde{L}}(\w^0)-\mathcal{\tilde{L}}(\w^*)}{\rho T}.
\end{equation}
Lemma~\ref{lemma:wellposed} shows that a solution optimizing $\tilde{\mathcal{L}}_{k}$ also satisfies $\mathcal{L}_{k}$. We finally get convergence rate of $O(\frac{1}{T})$ for \algoName.

\section{Conclusion}
In this paper, we propose \algoName, an effective and efficient FL algorithm designed to address the challenges of cross-domain scenarios and heterogeneous model architectures. By introducing Adjustable Teacher-Student Mutual Evaluation Curriculum Learning, \algoName significantly improves the effectiveness of knowledge distillation. Our experimental results demonstrate that \algoName outperforms state-of-the-art baselines in terms of both model performance and communication efficiency. Furthermore, we demonstrate the benefits of carefully designed models in enhancing FL outcomes in heterogeneous model architecture scenarios.
We analyze the algorithm that achieves a convergence rate of $O(\frac{1}{T})$.

\vfill\pagebreak


\bibliographystyle{IEEEbib}
\label{sec:refs}
\ninept
\bibliography{refs}

\begin{thebibliography}{10}

\bibitem{FedAvg}
Brendan McMahan, Eider Moore, Daniel Ramage, et~al.,
\newblock ``Communication-efficient learning of deep networks from
  decentralized data,''
\newblock in {\em {AISTATS}}. 2017, vol.~54 of {\em Proceedings of Machine
  Learning Research}, pp. 1273--1282, {PMLR}.

\bibitem{kairouzAdvancesOpenProblems2021a}
Peter Kairouz, H.~Brendan McMahan, Brendan Avent, et~al.,
\newblock ``Advances and open problems in federated learning,''
\newblock {\em Found. Trends Mach. Learn.}, vol. 14, no. 1-2, pp. 1--210, 2021.

\bibitem{FedMD}
Daliang Li et~al.,
\newblock ``Fedmd: Heterogenous federated learning via model distillation,''
\newblock {\em CoRR}, vol. abs/1910.03581, 2019.

\bibitem{tanPersonalizedFederatedLearning2022}
Alysa~Ziying Tan, Han Yu, Lizhen Cui, and Qiang Yang,
\newblock ``Towards {{Personalized Federated Learning}},''
\newblock {\em IEEE Transactions on Neural Networks and Learning Systems}, pp.
  1--17, 2022.

\bibitem{hintonDistillingKnowledgeNeural2015}
Geoffrey~E. Hinton, Oriol Vinyals, and Jeffrey Dean,
\newblock ``Distilling the knowledge in a neural network,''
\newblock {\em CoRR}, vol. abs/1503.02531, 2015.

\bibitem{FedRep}
Liam Collins, Hamed Hassani, Aryan Mokhtari, and Sanjay Shakkottai,
\newblock ``Exploiting shared representations for personalized federated
  learning,''
\newblock in {\em {ICML}}. 2021, vol. 139 of {\em Proceedings of Machine
  Learning Research}, pp. 2089--2099, {PMLR}.

\bibitem{FeDGen}
Zhuangdi Zhu, Junyuan Hong, and Jiayu Zhou,
\newblock ``Data-free knowledge distillation for heterogeneous federated
  learning,''
\newblock in {\em {ICML}}. 2021, vol. 139 of {\em Proceedings of Machine
  Learning Research}, pp. 12878--12889, {PMLR}.

\bibitem{FedRoD}
Hong-You Chen and Wei-Lun Chao,
\newblock ``On bridging generic and personalized federated learning for image
  classification,''
\newblock in {\em {ICLR}}, 2022.

\bibitem{CD2-pFed}
Yiqing Shen et~al.,
\newblock ``Cd2-pfed: Cyclic distillation-guided channel decoupling for model
  personalization in federated learning,''
\newblock in {\em {CVPR}}. 2022, pp. 10031--10040, {IEEE}.

\bibitem{FURL}
Duc Bui, Kshitiz Malik, Jack Goetz, Honglei Liu, Seungwhan Moon, Anuj Kumar,
  and Kang~G. Shin,
\newblock ``Federated user representation learning,''
\newblock {\em CoRR}, vol. abs/1909.12535, 2019.

\bibitem{FedKD}
Chuhan Wu, Fangzhao Wu, et~al.,
\newblock ``Communication-efficient federated learning via knowledge
  distillation,''
\newblock {\em Nature Communications}, vol. 13, no. 1, pp. 2032, Apr. 2022.

\bibitem{heGroupKnowledgeTransfer2020}
Chaoyang He, Murali Annavaram, and Salman Avestimehr,
\newblock ``Group knowledge transfer: Federated learning of large cnns at the
  edge,''
\newblock in {\em NeurIPS}, 2020.

\bibitem{makhijaArchitectureAgnosticFederated2022}
Disha Makhija, Xing Han, Nhat Ho, and Joydeep Ghosh,
\newblock ``Architecture agnostic federated learning for neural networks,''
\newblock in {\em {ICML}}. 2022, vol. 162 of {\em Proceedings of Machine
  Learning Research}, pp. 14860--14870, {PMLR}.

\bibitem{FedProto}
Yue Tan et~al.,
\newblock ``Fedproto: Federated prototype learning across heterogeneous
  clients,''
\newblock in {\em Proceedings of the AAAI Conference on Artificial
  Intelligence}, 2022, vol.~36, pp. 8432--8440.

\bibitem{FedPAC}
Jian Xu, Xinyi Tong, and Shao-Lun Huang,
\newblock ``Personalized federated learning with feature alignment and
  classifier collaboration,''
\newblock in {\em {ICLR}}, 2023.

\bibitem{FedPCL}
Yue Tan, Guodong Long, Jie Ma, Lu~Liu, Tianyi Zhou, and Jing Jiang,
\newblock ``Federated learning from pre-trained models: {A} contrastive
  learning approach,''
\newblock in {\em NeurIPS}, 2022.

\bibitem{resnet}
Kaiming He, Xiangyu Zhang, Shaoqing Ren, and Jian Sun,
\newblock ``Deep residual learning for image recognition,''
\newblock in {\em {CVPR}}. 2016, pp. 770--778, {IEEE} Computer Society.

\bibitem{CLSurvey}
Xin Wang, Yudong Chen, and Wenwu Zhu,
\newblock ``A survey on curriculum learning,''
\newblock {\em {IEEE} Trans. Pattern Anal. Mach. Intell.}, vol. 44, no. 9, pp.
  4555--4576, 2022.

\bibitem{wang2020attack}
Hongyi Wang et~al.,
\newblock ``Attack of the tails: Yes, you really can backdoor federated
  learning,''
\newblock {\em Advances in Neural Information Processing Systems}, vol. 33, pp.
  16070--16084, 2020.

\bibitem{FedBABU}
Jaehoon Oh, Sangmook Kim, and Se{-}Young Yun,
\newblock ``Fedbabu: Toward enhanced representation for federated image
  classification,''
\newblock in {\em {ICLR}}. 2022, OpenReview.net.

\bibitem{MNIST}
Yann LeCun, L{\'{e}}on Bottou, Yoshua Bengio, and Patrick Haffner,
\newblock ``Gradient-based learning applied to document recognition,''
\newblock {\em Proc. {IEEE}}, vol. 86, no. 11, pp. 2278--2324, 1998.

\bibitem{MNIST-M}
Yaroslav Ganin et~al.,
\newblock ``Unsupervised domain adaptation by backpropagation,''
\newblock in {\em {ICML}}. 2015, vol.~37 of {\em {JMLR} Workshop and Conference
  Proceedings}, pp. 1180--1189, JMLR.org.

\bibitem{mobilenet}
Andrew Howard, Ruoming Pang, et~al.,
\newblock ``Searching for mobilenetv3,''
\newblock in {\em {ICCV}}. 2019, pp. 1314--1324, {IEEE}.

\bibitem{vgg}
Karen Simonyan and Andrew Zisserman,
\newblock ``Very deep convolutional networks for large-scale image
  recognition,''
\newblock in {\em {ICLR}}, 2015.

\bibitem{UCI}
Dheeru Dua and Casey Graff,
\newblock ``{UCI} machine learning repository,'' 2017.

\bibitem{HCC}
J.~Best, H.~Bilgi, et~al.,
\newblock ``The {{GALAD}} scoring algorithm based on {{AFP}}, {{AFP-L3}}, and
  {{DCP}} significantly improves detection of {{BCLC}} early stage
  hepatocellular carcinoma,''
\newblock {\em Z Gastroenterol}, vol. 54, no. 12, pp. 1296--1305, Dec. 2016.

\bibitem{FedALA}
Jianqing Zhang, Yang Hua, Hao Wang, Tao Song, Zhengui Xue, et~al.,
\newblock ``Fedala: Adaptive local aggregation for personalized federated
  learning,''
\newblock in {\em Proceedings of the AAAI Conference on Artificial
  Intelligence}, 2023, vol.~37, pp. 11237--11244.

\bibitem{FedSSD}
Yuting He, Yiqiang Chen, XiaoDong Yang, et~al.,
\newblock ``Learning critically: Selective self-distillation in federated
  learning on non-iid data,''
\newblock {\em IEEE Transactions on Big Data}, 2022.

\bibitem{FedGKD}
Dezhong Yao, Wanning Pan, Yutong Dai, Yao Wan, Xiaofeng Ding, Hai Jin, Zheng
  Xu, and Lichao Sun,
\newblock ``Local-global knowledge distillation in heterogeneous federated
  learning with non-iid data,''
\newblock {\em CoRR}, vol. abs/2107.00051, 2021.

\bibitem{FedProx}
Tian Li, Anit~Kumar Sahu, Manzil Zaheer, Maziar Sanjabi, Ameet Talwalkar, and
  Virginia Smith,
\newblock ``Federated optimization in heterogeneous networks,''
\newblock in {\em MLSys}. 2020, mlsys.org.

\end{thebibliography}

\end{document}